\documentclass[11pt,a4paper]{article}
\usepackage{authblk}
\usepackage[hyperref]{naaclhlt2019}
\usepackage{times}
\usepackage{latexsym}
\usepackage{url}
\usepackage{graphicx}

\aclfinalcopy % Uncomment this line for the final submission

\title{Fast Prototyping a Dialogue Comprehension System for Nurse-Patient \\Conversations on Symptom Monitoring}

\author[$\star$]{\textbf{Zhengyuan Liu}}
\author[$\star$]{\textbf{Hazel Lim}}
\author[$\star$]{\textbf{Nur Farah Ain Binte Suhaimi}}
\author[$\dagger$]{\textbf{Shao Chuen Tong}}
\author[$\ddagger$]{\\ \textbf{Sharon Ong}}
\author[$\ddagger$]{\textbf{Angela Ng}}
\author[$\top$]{\textbf{Sheldon Lee}}
\author[$\top$]{\textbf{Michael R. Macdonald}}
\author[$\star$]{\\ \textbf{Savitha Ramasamy}}
\author[$\star$]{\textbf{Pavitra Krishnaswamy}}
\author[$\dagger$]{\textbf{Wai Leng Chow}}
\author[$\star$]{\textbf{Nancy F. Chen}}

\affil[$\star$]{Institute for Infocomm Research, A*STAR, Singapore}
\affil[$\dagger$]{Health Services Research, Changi General Hospital, Singapore}
\affil[$\ddagger$]{Health Management Unit, Changi General Hospital, Singapore}
\affil[$\top$]{Department of Cardiology, Changi General Hospital, Singapore}  
\affil[ ]{\tt{\{liu\_zhengyuan, nfychen\}@i2r.a-star.edu.sg}}

\date{}

\begin{document}
\maketitle
\begin{abstract}
Data for human-human spoken dialogues for research and development are currently very limited in quantity, variety, and sources; such data are even scarcer in healthcare. In this work, we investigate fast prototyping of a dialogue comprehension system by leveraging on minimal nurse-to-patient conversations. We propose a framework inspired by nurse-initiated clinical symptom monitoring conversations to construct a simulated human-human dialogue dataset, embodying linguistic characteristics of spoken interactions like thinking aloud, self-contradiction, and topic drift. We then adopt an established bidirectional attention pointer network on this simulated dataset, achieving more than 80\% F1 score on a held-out test set from real-world nurse-to-patient conversations. The ability to automatically comprehend conversations in the healthcare domain by exploiting only limited data has implications for improving clinical workflows through red flag symptom detection and triaging capabilities. We demonstrate the feasibility for efficient and effective extraction, retrieval and comprehension of symptom checking information discussed in multi-turn human-human spoken conversations.
\end{abstract}

\section{Introduction}
\subsection{Problem Statement}
Spoken conversations still remain the most natural and effortless means of human communication. Thus a lot of valuable information is conveyed and exchanged in such an unstructured form. In telehealth settings, nurses might call discharged patients who have returned home to continue to monitor their health status. Human language technology that can efficiently and effectively extract key information from such conversations is clinically useful, as it can help streamline workflow processes and digitally document patient medical information to increase staff productivity.
In this work, we design and prototype a dialogue comprehension system in the question-answering manner, which is able to comprehend spoken conversations between nurses and patients to extract clinical information\footnote{N.F.C., P.K., R.S. and C.W.L. conceptualized the overall research programme; N.F.C. and L.Z. proposed and developed the proposed approach; L.Z. and N.F.C. developed methods for data analysis; L.Z., L.J.H., N.F.S., and N.F.C. constructed the corpus; T.S.C, S.O., A.N. S.L.G, and M.R.M acquired, prepared, and validated clinical data; L.Z., N.F. C., P.K., S.L.G., M.R.M and C.W.L interpreted results; L.Z., N.F.C., L.J.H, N.F.S., P.K. and C.W.L wrote the paper.}.

\subsection{Motivation of Approach}
Machine comprehension of written passages has made tremendous progress recently. Large quantities of supervised training data for reading comprehension (e.g. SQuAD \cite{rajpurkar2016squad_1}), the wide adoption and intense experimentation of neural modeling \cite{seo2016bidaf, wang2017rnet}, and the advancements in vector representations of word embeddings \cite{pennington2014glove,devlin2018bert} all contribute significantly to the achievements obtained so far. The first factor, the availability of large scale datasets, empowers the latter two factors. 
To date, there is still very limited well-annotated large-scale data suitable for modeling human-human spoken dialogues. Therefore, it is not straightforward to directly port over the recent endeavors in reading comprehension to dialogue comprehension tasks.

In healthcare, conversation data is even scarcer due to privacy issues. Crowd-sourcing is an efficient way to annotate large quantities of data, but less suitable for healthcare scenarios, where domain knowledge is required to guarantee data quality. To demonstrate the feasibility of a dialogue comprehension system used for extracting key clinical information from symptom monitoring conversations, we developed a framework to construct a simulated human-human dialogue dataset to bootstrap such a prototype. Similar efforts have been conducted for human-machine dialogues for restaurant or movie reservations \cite{shah2018dialogueself}. To the best of our knowledge, no one to date has done so for human-human conversations in healthcare.

\subsection{Human-human Spoken Conversations}
\begin{figure}[t]
\centering
\includegraphics[width=7.7cm]{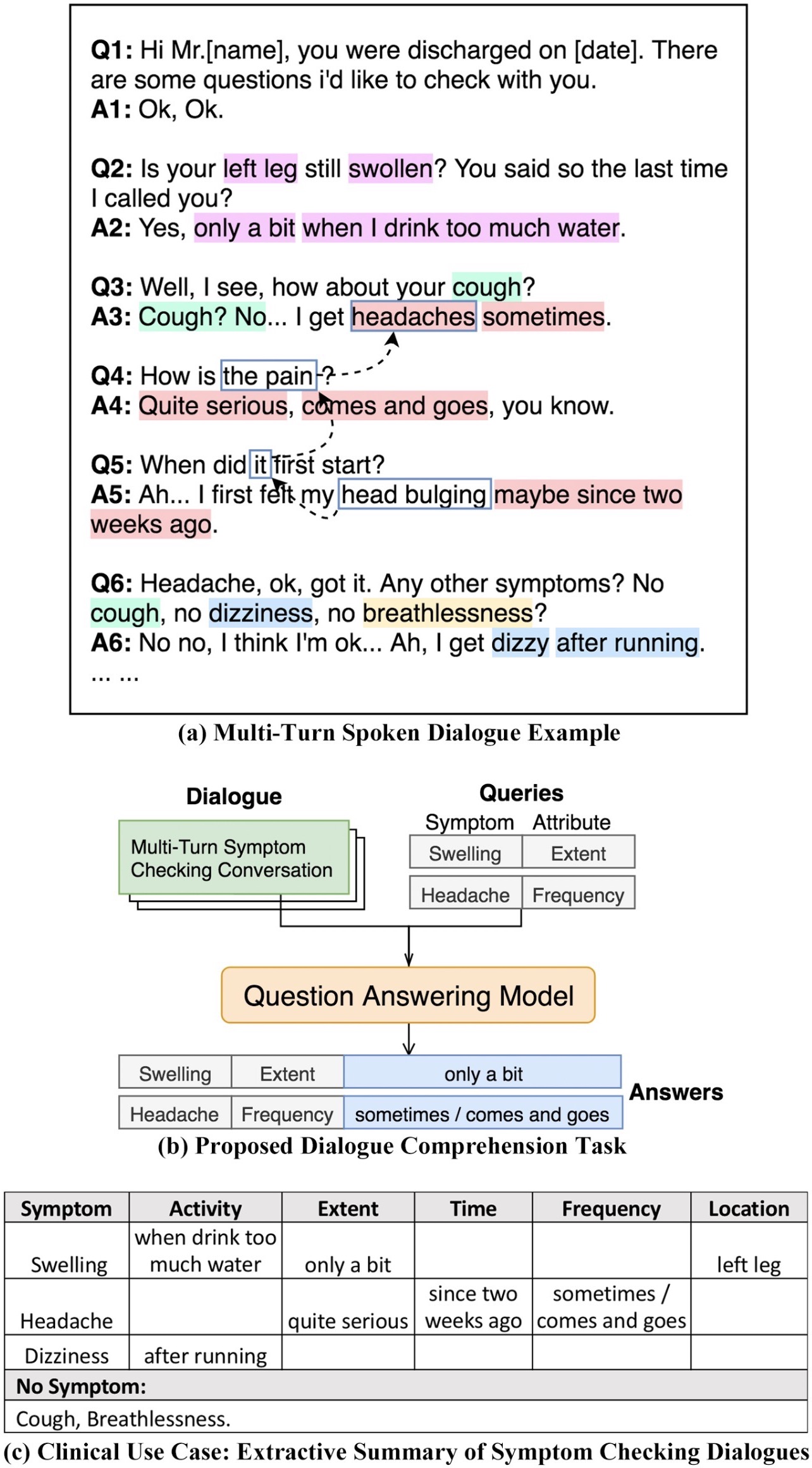}
\caption{Dialogue comprehension of symptom checking conversations.}
\label{system-fig}
\vspace{-0.5cm}
\end{figure}

Human-human spoken conversations are a dynamic and interactive flow of information exchange. While developing technology to comprehend such spoken conversations presents similar technical challenges as machine comprehension of written passages \cite{rajpurkar2018squad_2}, the challenges are further complicated by the interactive nature of human-human spoken conversations: 

(1) Zero anaphora is more common: Co-reference resolution of spoken utterances from multiple speakers is needed. For example, in Figure \ref{system-fig}(a) \textit{headaches, the pain, it, head bulging} all refer to the patient's headache symptom, but they were uttered by different speakers and across multiple utterances and turns. In addition, anaphors are more likely to be omitted (see Figure \ref{system-fig}(a) A4) as this does not affect the human listener’s understanding, but it might be challenging for computational models. 

(2) Thinking aloud more commonly occurs: Since it is more effortless to speak than to type, one is more likely to reveal her running thoughts when talking. In addition, one cannot retract what has been uttered, while in text communications, one is more likely to confirm the accuracy of the information in a written response and revise if necessary before sending it out. 
Thinking aloud can lead to self-contradiction, requiring more context to fully understand the dialogue; e.g., in A6 in Figure \ref{system-fig}(a), the patient at first says he has none of the symptoms asked, but later revises his response saying that he does get dizzy after running.

(3) Topic drift is more common and harder to detect in spoken conversations: An example is shown in Figure \ref{system-fig}(a) in A3, where \textit{No} is actually referring to \textit{cough} in the previous question, and then the topic is shifted to \textit{headache}. 
In spoken conversations, utterances are often incomplete sentences so traditional linguistic features used in written passages such as punctuation marks indicating syntactic boundaries or conjunction words suggesting discourse relations might no longer exist. 

\subsection{Dialogue Comprehension Task}
\label{task-definition}
Figure~\ref{system-fig}(b) illustrates the proposed dialogue comprehension task using a question answering (QA) model. The input are a multi-turn symptom checking dialogue $D$ and a query $Q$ specifying a \textit{symptom} with one of its \textit{attributes}; the output is the extracted answer $A$ from the given dialogue. A training or test sample is defined as $S=\{D, Q,A\}$. Five attributes, specifying certain details of clinical significance, are defined to characterize the answer types of $A$: (1) the \textit{time} the patient has been experiencing the symptom, (2) \textit{activities} that trigger the symptom (to occur or worsen), (3) the \textit{extent} of seriousness, (4) the \textit{frequency} occurrence of the symptom, and (5) the \textit{location} of symptom.  For each symptom/attribute, it can take on different linguistic expressions, defined as \textit{entities}. Note that if the queried symptom or attribute is not mentioned in the dialogue, the groundtruth output is ``No Answer'', as in \cite{rajpurkar2018squad_2}.

\section{Related Work}
\subsection{Reading Comprehension}
Large-scale reading comprehension tasks like SQuAD \cite{rajpurkar2016squad_1} and MARCO \cite{nguyen2016msmarco} provide question-answer pairs from a vast range of written passages, covering different kinds of factual answers involving entities such as location and numerical values. Furthermore, HotpotQA \cite{yang2018hotpotqa} requires multi-step inference and provides numerous answer types. CoQA \cite{reddy2018coqa} and QuAC \cite{choi2018quac} are designed to mimic multi-turn information-seeking discussions of the given material. In these tasks, contextual reasoning like coreference resolution is necessary to grasp rich linguistic patterns, encouraging semantic modeling beyond naive lexical matching. Neural networks contribute to impressive progress in semantic modeling: distributional semantic word embeddings \cite{pennington2014glove}, contextual sequence encoding \cite{sutskever2014seq2seq,gehring2017fairseq} and the attention mechanism \cite{luong2015attention, vaswani2017transformer} are widely adopted in state-of-the-art comprehension models \cite{seo2016bidaf,wang2017rnet,devlin2018bert}.

While language understanding tasks in dialogue such as domain identification \cite{ravuri2015utter_cls}, slot filling \cite{kurata2016slotfilling} and user intent detection \cite{wen2016woz_2} have attracted much research interest, work in dialogue comprehension is still limited, if any.  
It is labor-intensive and time-consuming to obtain a critical mass of annotated conversation data for computational modeling. Some propose to collect text data from human-machine or machine-machine dialogues \cite{li2016usersimulator,shah2018dialogueself}. In such cases, as human speakers are aware of current limitations of dialogue systems or due to pre-defined assumptions of user simulators, there are fewer cases of zero anaphora, thinking aloud, and topic drift, which occur more often in human-human spoken interactions. 

\subsection{NLP for Healthcare}
There is emerging interest in research and development activities at the intersection of machine learning and healthcare\footnote{ML4H: Machine Learning for Health, Workshop at NeurIPS 2018 https://ml4health.github.io/2018/}
\footnote{2018 Workshop on Health Intelligence (W3PHIAI 2018) http://w3phiai2018.w3phi.com/}, of which much of the NLP related work are centered around social media or online forums (e.g., \cite{wallace2014onlinereview, lyles2013cancer_twitter}), partially due to the world wide web as a readily available source of information.  
Other work in this area uses public data sources such as MIMIC\footnote{https://mimic.physionet.org/} in electronic health records: text classification approaches have been applied to analyze unstructured clinical notes for ICD code assignment \cite{baumel2017icd_classification} and automatic intensive emergency prediction \cite{grnarova2016ehr_notes}. Sequence-to-sequence textual generation has been used for readable notes based on medical and demographic recordings \cite{liu2018write_notes}. 
 For mental health, there has been more focus on analyzing dialogues. For example, 
 sequential modeling of audio and text have helped detect depression from human-machine interviews  \cite{al2018detectdepression}. However, few studies have examined human-human spoken conversations in healthcare settings.

\section{Real-World Data Analysis}
\subsection{Data Preparation}
We used recordings of nurse-initiated telephone conversations for congestive heart failure patients undergoing telemonitoring, post-discharge from the hospital. The clinical data was acquired by the Health Management Unit at Changi General Hospital. This research study was approved by the SingHealth Centralised Institutional Review Board (Protocol 1556561515). The patients were recruited during 2014-2016 as part of their routine care delivery, and enrolled into the telemonitoring health management program with consent for use of anonymized versions of their data for research.

The dataset comprises a total of 353 conversations from 40 speakers (11 nurses, 16 patients, and 13 caregivers) with consent to the use of anonymized data for research. The speakers are 38 to 88 years old, equally distributed across gender, and comprise a range of ethnic groups (55\% Chinese, 17\% Malay, 14\% Indian, 3\% Eurasian, and 11\% unspecified).
The conversations cover 11 topics (e.g., medication compliance, symptom checking, education, greeting) and 9 symptoms (e.g., chest pain, cough) and amount to 41 hours.

Data preprocessing and anonymization were performed by a data preparation team, separate from the data analysis team to maintain data confidentiality. The data preparation team followed standard speech recognition transcription guidelines, where words are transcribed verbatim to include false starts, disfluencies, mispronunciations, and private self-talk. Confidential information were marked and clipped off from the audio and transcribed with predefined tags in the annotation.  Conversation topics and clinical symptoms were also annotated and clinically validated by certified telehealth nurses.

\subsection{Linguistic Characterization on Seed Data} \label{sec:seed}
To analyze the linguistic structure of the inquiry-response pairs in the entire 41-hour dataset, we randomly sampled a seed dataset consisting of 1,200 turns and manually categorized them to different types, which are summarized in Table~\ref{utterance-table} along with the corresponding occurrence frequency statistics. Note that each given utterance could be categorized to more than one type. We elaborate on each utterance type below.

\begin{table}[t]
\begin{center}
\small
\begin{tabular}{p{6cm}r}
\hline \bf Utterance Type & \bf \% \\ \hline
Open-ended Inquiry & 31.8 \\
Detailed Inquiry & 33.0 \\
Multi-Intent Inquiry & 15.5 \\
Reconfirmation Inquiry & 21.3 \\
Inquiry with Transitional Clauses & 8.5 \\
\hline
Yes/No Response & 52.1 \\
Detailed Response & 29.4 \\
Response with Revision & 5.1 \\
Response with Topic Drift & 11.1 \\
Response with Transitional Clauses & 9.5 \\
\hline
\bf Sampled Turn Number & 1200 \\
\hline
\end{tabular}
\end{center}
\caption{\label{utterance-table} Linguistic characterization of inquiry-response types and their occurrence frequency from the seed data in Section \ref{sec:seed}.
}
\end{table}

\noindent{\bf Open-ended Inquiry:}
Inquiries about general well-being or a particular symptom; e.g., \textit{``How are you feeling?''} and \textit{``Do you cough?''}\\
\noindent{\bf Detailed Inquiry:}
Inquiries with specific details that prompt yes/no answers or clarifications; e.g., \textit{``Do you cough at night?''}\\
\noindent{\bf Multi-Intent Inquiry:}
Inquiring more than one symptom in a question; e.g., \textit{``Any cough, chest pain, or headache?''}\\
\noindent{\bf Reconfirmation Inquiry:}
The nurse reconfirms particular details; e.g., \textit{``Really? At night?''} and \textit{``Serious or mild?''}. This case is usually related to explicit or implicit coreferencing. \\
\noindent{\bf Inquiry with Transitional Clauses:}
During spoken conversations, one might repeat what the other party said, but it is unrelated to the main clause of the question. This is usually due to private self-talk while thinking aloud, and such utterances form a transitional clause before the speaker starts a new topic; e.g.,  \textit{``Chest pain... no chest pain, I see... any cough?''}.\\
\noindent{\bf Yes/No Response:}
Yes/No responses seem straightforward, but sometimes lead to misunderstanding if one does not interpret the context appropriately. One case is tag questions: \textit{A:``You don't cough at night, do you?'' B:`Yes, yes'' A:``cough at night?'' B:``No, no cough''}.
Usually when the answer is unclear, clarifying inquiries will be asked for reconfirmation purposes. 
\\
\noindent{\bf Detailed Response:}
Responses that contain specific information of one symptom, like \textit{``I felt tightness in my chest''}.\\
\noindent{\bf Response with Revision:}
Revision is infrequent but can affect comprehension significantly. One cause is thinking aloud so a later response overrules the previous one; e.g., \textit{``No dizziness, oh wait... last week I felt a bit dizzy when biking''}.\\
\noindent{\bf Response with Topic Drift:}
When a symptom/topic like headache is inquired, the response might be: \textit{``Only some chest pain at night''}, not referring to the original symptom (headache) at all.\\
\noindent{\bf Response with Transitional Clauses:}
Repeating some of the previous content, but often unrelated to critical clinical information and usually followed by topic drift. For example, \textit{``Swelling... swelling... I don't cough at night''}.

\begin{figure*}[htp]
\centering
\includegraphics[width=16cm]{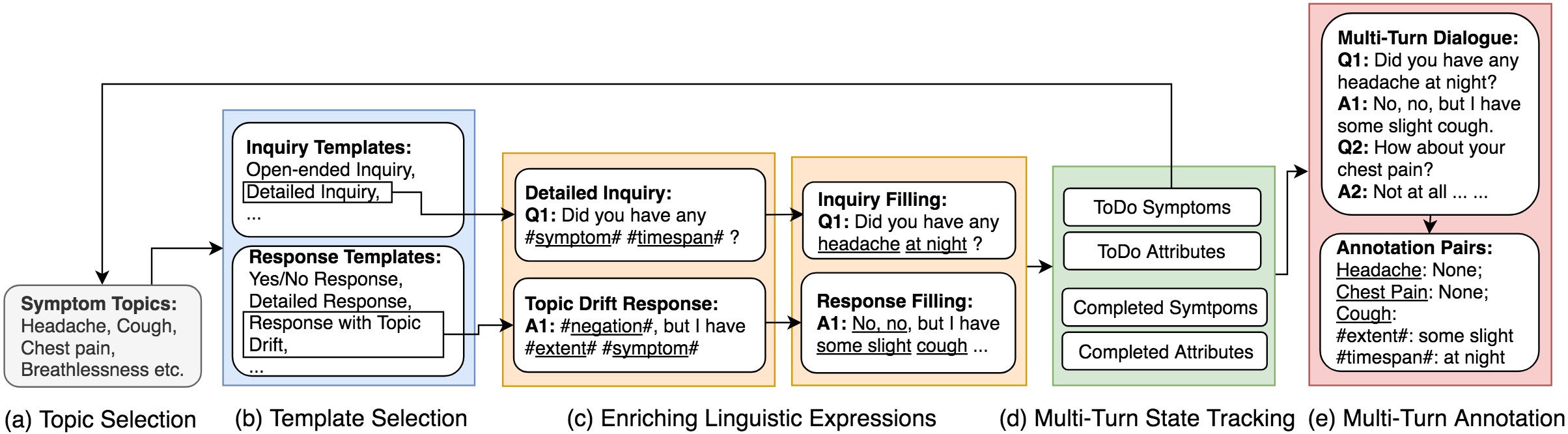}
\caption{Simulated data generation framework.}
\label{generator-fig}
\vspace{-0.3cm}
\end{figure*}

\section{Simulating Symptom Monitoring Dataset for Training}
We divide the construction of data simulation into two stages. In Section \ref{sec:template}, we build templates and expression pools using linguistic analysis followed by manual verification. In Section \ref{sec:framework}, we present our proposed framework for generating simulated training data. The templates and framework are verified for logical correctness and clinical soundness.

\subsection{Template Construction}\label{sec:template}
\subsubsection{Linguistically-Inspired Templates}
Each utterance in the seed data is categorized according to Table \ref{utterance-table} and then abstracted into templates by replacing entity phrases like \textit{cough} and \textit{often} with respective placeholders \textit{``\#symptom\#''} and \textit{``\#frequency\#''}. The templates are refined through verifying logical correctness and injecting expression diversity by linguistically trained researchers.  
As these replacements do not alter the syntactic structure, we interchange such placeholders with various verbal expressions to enlarge the simulated training set in Section \ref{sec:framework}. Clinical validation was also conducted by certified telehealth nurses.

\subsubsection{Topic Expansion \& Symptom Customization}
For the 9 symptoms (e.g. chest pain, cough) and 5 attributes (e.g., extent, frequency), we collect various expressions from the seed data, and expand them through synonym replacement. Some attributes are unique to a particular symptom; e.g., \textit{``left leg''} in \textit{\#location\#} is only suitable to describe the symptom \textit{swelling}, but not the symptom \textit{headache}. Therefore, we only reuse general expressions like \textit{``slight''} in \textit{\#extent\#} across different symptoms to diversify linguistic expressions.

\subsubsection{Expression Pool for Linguistic Diversity}\label{sec:pool}
Two linguistically trained researchers constructed expression pools for each symptom and each attribute to account for different types of paraphrasing and descriptions. These expression pools are used in Section \ref{sec:framework} (c). 

\subsection{Simulated Data Generation Framework}\label{sec:framework}
 Figure~\ref{generator-fig} shows the five steps we use to generate multi-turn symptom monitoring dialogue samples.

\noindent{\bf (a) Topic Selection:}
While nurses might prefer to inquire the symptoms in different orders depending on the patient's history, our preliminary analysis shows that modeling results do not differ noticeably if topics are of equal prior probabilities. Thus we adopt this assumption for simplicity.  

\noindent{\bf (b) Template Selection:}
For each selected topic, one inquiry template and one response template are randomly chosen to compose a turn. 
To minimize adverse effects of underfitting, we redistributed the frequency distribution in Table \ref{utterance-table}: For utterance types that are below 15\%, we boosted them to 15\%, and the overall relative distribution ranking is balanced and consistent with Table \ref{utterance-table}.

\noindent{\bf (c) Enriching Linguistic Expressions:}
The placeholders in the selected templates are substituted with diverse expressions from the expression pools in Section \ref{sec:pool} to characterize the symptoms and their corresponding attributes.

\noindent{\bf (d) Multi-Turn Dialogue State Tracking:}
A greedy algorithm is applied to complete conversations. A ``completed symptoms'' list and a ``to-do symptoms'' list are used for symptom topic tracking. We also track the ``completed attributes" and ``to-do attributes". For each symptom, all related attributes are iterated. A dialogue ends only when all possible entities are exhausted, generating a multi-turn dialogue sample, which encourages the model to learn from the entire discussion flow rather than a single turn to comprehend contextual dependency. The average length of a simulated dialogue is 184 words, which happens to be twice as long as an average dialogue from the real-world evaluation set. Moreover, to model the roles of the respondents, we set the ratio between patients and caregivers to 2:1; this statistic is inspired by the real scenarios in the seed dataset. For both the caregivers and patients, we assume equal probability of both genders. The corresponding pronouns in the conversations are thus determined by the role and gender of these settings. 

\noindent{\bf (e) Multi-Turn Sample Annotation:}
For each multi-turn dialogue, a query is specified by a symptom and an attribute. The groundtruth output of the QA system is automatically labeled based on the template generation rules, but also manually verified to ensure annotation quality. Moreover, we adopt the unanswerable design in \cite{rajpurkar2018squad_2}: when the patient does not mention a particular symptom, the answer is defined as \textit{``No Answer''}. This process is repeated until all logical permutations of symptoms and attributes are exhausted.

\begin{figure}[htp]
\centering
\includegraphics[width=7.7cm]{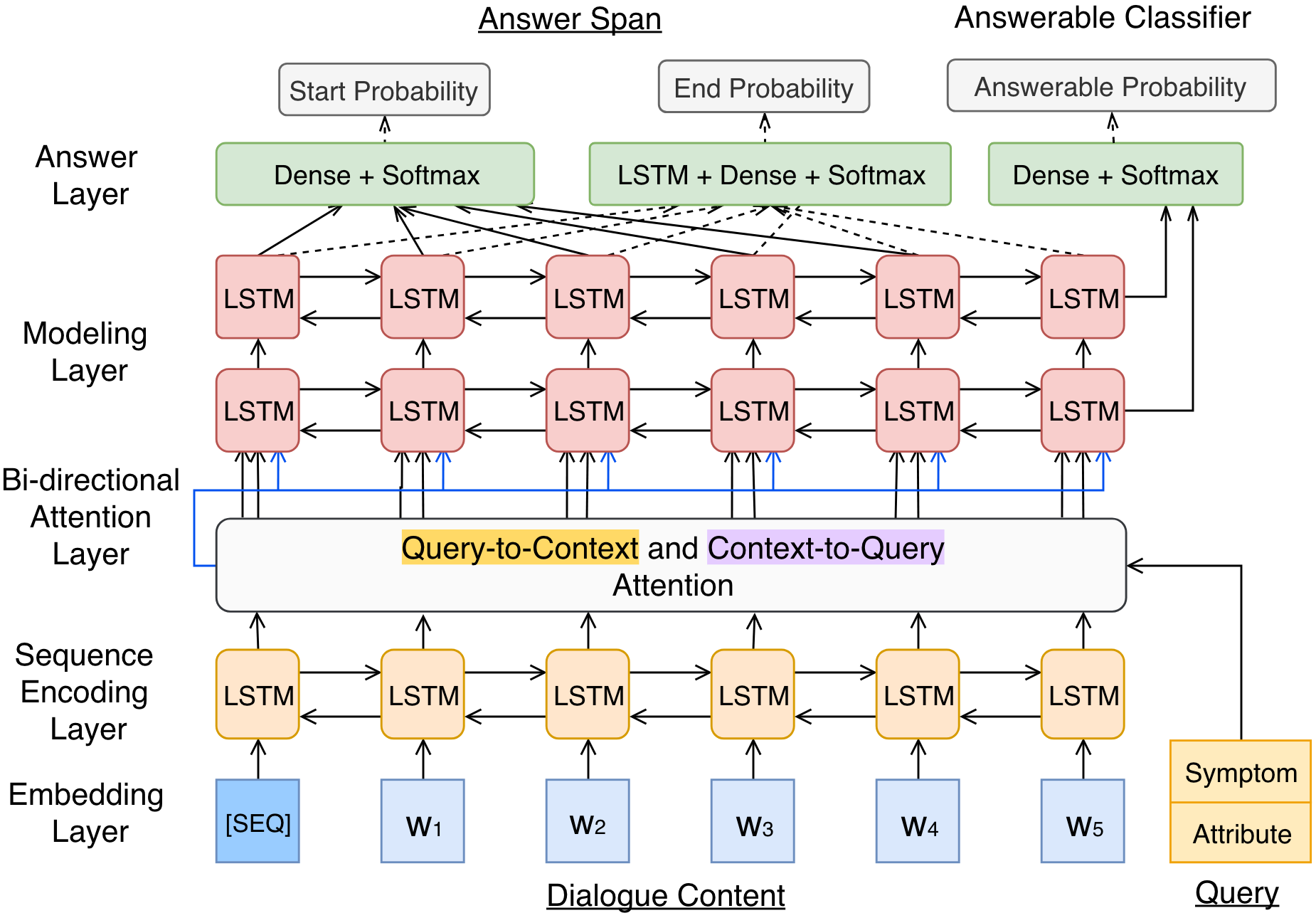}
\caption{Bi-directional attention pointer network with an answerable classifier for dialogue comprehension.}
\label{biattention-fig}
\vspace{-0.5cm}
\end{figure}

\section{Experiments}\vspace{-0.cm}
\subsection{Model Design}\vspace{-0.cm}
We implemented an established model in reading comprehension, a bi-directional attention pointer network \cite{seo2016bidaf}, and equipped it with an answerable classifier, as depicted in Figure~\ref{biattention-fig}. First, tokens in the given dialogue $D$ and query $Q$ are converted into embedding vectors. Then the dialogue embeddings are fed to a bi-directional LSTM encoding layer, generating a sequence of contextual hidden states. Next, the hidden states and query embeddings are processed by a bi-directional attention layer, fusing attention information in both context-to-query and query-to-context directions. The following two bi-directional LSTM modeling layers read the contextual sequence with attention. Finally, two respective linear layers with softmax functions are used to estimate token $i$'s $p_i^{start}$ and $p_i^{end}$ probability of the answer span $A$. \\
In addition, we add a special tag ``[SEQ]'' at the head of $D$ to account for the case of  \textit{``No answer''} \cite{devlin2018bert} and adopt an answerable classifier as in  \cite{liu2018verifierSAN}. More specifically, when the queried symptom or attribute is not mentioned in the dialogue, the answer span should point to the tag ``[SEQ]'' and answerable probability should be predicted as 0.

\subsection{Implementation Details}
The model was trained via gradient backpropagation with the cross-entropy loss function of answer span prediction and answerable classification, optimized by Adam algorithm \cite{kingma2014adam} with initial learning rate of $0.001$.  Pre-trained GloVe \cite{pennington2014glove} embeddings (size $=200$) were used. We re-shuffled training samples at each epoch (batch size $=16$). Out-of-vocabulary words ($< 0.05\%$) were replaced with a fixed random vector. L2 regularization and dropout (rate $=0.2$) were used to alleviate overfitting \cite{srivastava2014dropout}.

\subsection{Evaluation Setup}
To evaluate the effectiveness of our linguistically-inspired simulation approach, the model is trained on the simulated data (see Section \ref{sec:framework}). We designed 3 evaluation sets: (1) \textbf{Base Set} (1,264 samples) held out from the simulated data. (2) \textbf{Augmented Set} (1,280 samples) built by adding two out-of-distribution symptoms, with corresponding dialogue contents and  queries, to the Base Set (\textit{``bleeding''} and \textit{``cold''}, which never appeared in training data). (3) \textbf{Real-World Set} (944 samples) manually delineated from the the symptom checking portions (approximately 4 hours) of real-world dialogues, and annotated as evaluation samples.

\begin{table}[t!]
\begin{center}
\small
\begin{tabular}{lp{0.8cm}p{0.8cm}p{0.8cm}p{0.8cm}}
\hline \bf Training Samples & \bf 10k & \bf 50k & \bf 100k & \bf 150k \\ \hline
\hline
\multicolumn{5}{l}{\bf Base Evaluation Set:} \\
EM Score & 15.41 & 80.33 & 89.68 & \textbf{91.45} \\
F1 Score & 50.63 & 89.18 & 92.27 & \textbf{94.17} \\
\hline
\multicolumn{5}{l}{\bf Augmented Evaluation Set:} \\
EM Score & 11.59 & 57.22 & \textbf{78.29} & 72.12 \\
F1 Score & 49.36 & 74.53 & \textbf{85.69} & 82.75 \\
\hline
\multicolumn{4}{l}{\bf Real-World Evaluation Set:} \\
EM Score & 38.81 & 42.93 & \textbf{78.23} & 75.41 \\
F1 Score & 46.29 & 52.68 & \textbf{80.18} & 78.09 \\
\hline
\end{tabular}
\end{center}
\caption{\label{eval-result-table} QA model evaluation results. Each sample is a simulated multi-turn conversation.}
\vspace{-0.5cm}
\end{table}

\subsection{Results}
Evaluation results are in Table~\ref{eval-result-table} with exact match (EM) and F1 score in \cite{rajpurkar2016squad_1} metrics. To distinguish the correct answer span from the plausible ones which contain the same words, we measure the scores on the position indices of tokens. Our results show that both EM and F1 score increase with training sample size growing and the optimal size in our setting is 100k. The best-trained model performs well on both the Base Set and the Augmented Set, indicating that out-of-distribution symptoms do not affect the comprehension of existing symptoms and outputs reasonable answers for both in- and out-of-distribution symptoms. On the Real-World Set, we obtained 78.23 EM score and 80.18 F1 score respectively.

Error analysis suggests the performance drop from the simulated test sets is due to the following: 1) sparsity issues resulting from the expression pools excluding various valid but sporadic expressions. 2) nurses and patients occasionally chit-chat in the Real-World Set, which is not simulated in the training set. At times, these chit-chats make the conversations overly lengthy, causing the information density to be lower. These issues could potentially distract and confuse the comprehension model.
3) an interesting type of infrequent error source, caused by patients elaborating on possible causal relations of two symptoms. For example, a patient might say ``\textit{My giddiness may be due to all this cough}''.
We are currently investigating how to close this performance gap efficiently.

\begin{table}[t!]
\begin{center}
\small
\begin{tabular}{lp{1cm}p{1cm}}
\hline  & \bf EM & \bf F1 \\
\hline
\multicolumn{3}{l}{\bf Augmented Evaluation Set:} \\
Best-trained Model & \textbf{78.29} & \textbf{85.69} \\
w/o Bi-Attention & 72.08 & 78.57 \\
w/o Pre-trained Embedding & 56.98 & 72.31 \\
\hline
\multicolumn{3}{l}{\bf Real-World Evaluation Set:} \\
Best-trained Model & \textbf{78.23} & \textbf{80.18} \\
w/o Bi-Attention & 70.52 & 74.09 \\
w/o Pre-trained Embedding & 60.88 & 66.47 \\
\hline
\end{tabular}
\end{center}
\caption{\label{ablation-table} Ablation experiments on 100K training size.}
\vspace{-0.3cm}
\end{table}

\subsection{Ablation Analysis} 
To assess the effectiveness of bi-directional attention, we bypassed the bi-attention layer by directly feeding the contextual hidden states and query embeddings to the modeling layer.
To evaluate the pre-trained GloVe embeddings, we randomly initialized and trained the embeddings from scratch. These two procedures lead to 10\% and 18\% performance degradation on the Augmented Set and Real-World Set, respectively (see Table~\ref{ablation-table}).

\section{Conclusion}
We formulated a dialogue comprehension task motivated by the need in telehealth settings to extract key clinical information from spoken conversations between nurses and patients. We analyzed linguistic characteristics of real-world human-human symptom checking dialogues, constructed a simulated dataset based on linguistically inspired and clinically validated templates, and prototyped a QA system. The model works effectively on a simulated test set using symptoms excluded during training and on real-world conversations between nurses and patients. 
We are currently improving the model's dialogue comprehension capability in complex reasoning and context understanding and also applying the QA model to summarization and virtual nurse applications.

\section*{Acknowledgements}
Research efforts were supported by funding for Digital Health and Deep Learning I2R (DL2 SSF Project No: A1718g0045) and the Science and Engineering Research Council (SERC Project No: A1818g0044), A*STAR, Singapore. In addition, this work was conducted using resources and infrastructure provided by the Human Language Technology unit at I2R. The telehealth data acquisition was funded by the Economic Development Board (EDB), Singapore Living Lab Fund and Philips Electronics – Hospital to Home Pilot Project (EDB grant reference number: S14-1035-RF-LLF H and W).

We acknowledge valuable support and assistance from Yulong Lin, Yuanxin Xiang, and Ai Ti Aw at the Institute for Infocomm Research (I2R); Weiliang Huang at the Changi General Hospital (CGH) Department of Cardiology, Hong Choon Oh at CGH Health Services Research, and Edris Atikah Bte Ahmad, Chiu Yan Ang, and Mei Foon Yap of the CGH Integrated Care Department. 

We also thank Eduard Hovy and Bonnie Webber for insightful discussions and the anonymous reviewers for their precious feedback to help improve and extend this piece of work.

\bibliography{naaclhlt2019}
\bibliographystyle{acl_natbib}

\appendix

\end{document}